\documentclass{article}

\usepackage{arxiv}

\usepackage[utf8]{inputenc} 
\usepackage[T1]{fontenc}    
\usepackage{hyperref}       
\usepackage{url}            
\usepackage{booktabs}       
\usepackage{amsfonts}       
\usepackage{nicefrac}       
\usepackage{graphicx}
\usepackage{doi}
\usepackage{linguex}
\usepackage{colortbl}
\usepackage{tabulary}
\usepackage{xstring}
\usepackage{xcolor}
\usepackage{float}

\title{Ethos and Pathos in Online Group Discussions: \\Corpora for Polarisation Issues in Social Media}

\author{ \href{https://orcid.org/0009-0006-6012-4787}{\includegraphics[scale=0.06]{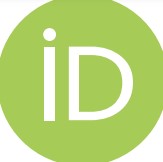}\hspace{1mm}Ewelina ~Gajewska} \\
	Laboratory of The New Ethos\\
	Warsaw University of Technology, Poland \\
	\texttt{ewelina.gajewska.dokt@pw.edu.pl} \\
	\And
	\href{https://orcid.org/0000-0001-9674-9902}{\includegraphics[scale=0.06]{orcid.jpg}\hspace{1mm}Katarzyna ~Budzynska} \\
	Laboratory of The New Ethos\\
	Warsaw University of Technology, Poland \\
	\texttt{katarzyna.budzynska@pw.edu.pl} \\
 	\And
	\href{https://orcid.org/0000-0003-2370-4636}{\includegraphics[scale=0.06]{orcid.jpg}\hspace{1mm}Barbara ~Konat} \\
	Department of Psychology and Cognitive Science\\
	Adam Mickiewicz University, Poland\\
	\texttt{barbara.konat@amu.edu.pl} \\
	\And
	\href{https://orcid.org/0000-0001-5553-7428}{\includegraphics[scale=0.06]{orcid.jpg}\hspace{1mm}Marcin ~Koszowy} \\
	Laboratory of The New Ethos\\
	Warsaw University of Technology, Poland \\
	\texttt{marcin.koszowy@pw.edu.pl} \\
	\And
	\href{https://orcid.org/0000-0003-1088-683}{\includegraphics[scale=0.06]{orcid.jpg}\hspace{1mm}Konrad ~Kiljan} \\
	Laboratory of Media Anthropology\\
	University of Warsaw, Poland\\
	\texttt{konrad.kiljan@uw.edu.pl} \\
 	\And
	\href{https://orcid.org/0009-0006-8953-8270}{\includegraphics[scale=0.06]{orcid.jpg}\hspace{1mm}Maciej ~Uberna} \\
	Laboratory of The New Ethos\\
	Warsaw University of Technology, Poland\\
	\texttt{maciej.uberna.dokt@pw.edu.pl} \\
 	\And
	\href{https://orcid.org/0009-0003-6610-2675}{\includegraphics[scale=0.06]{orcid.jpg}\hspace{1mm}He ~Zhang} \\
	Department of Administration and Social Sciences\\
	Warsaw University of Technology, Poland \\
	\texttt{he.zhang@pw.edu.pl}}
 
\renewcommand{\headeright}{}
\renewcommand{\undertitle}{}
\renewcommand{\shorttitle}{Corpora for Polarisation Issues}

\hypersetup{
pdftitle={Ethos and Pathos in Online Group Discussions: Corpora for Polarisation Issues in Social Media},
pdfsubject={ },
pdfauthor={Ewelina ~Gajewska},
pdfkeywords={rhetorical strategies, polarisation in society, multi-topic and multi-platform corpora},
}

\begin{document}
\maketitle

\begin{abstract}
Growing polarisation in society caught the attention of the scientific community as well as news media, which devote special issues to this phenomenon. At the same time, digitalisation of social interactions requires to revise concepts from social science regarding establishment of trust, which is a key feature of all human interactions, and group polarisation, as well as new computational tools to process large quantities of available data. Existing methods seem insufficient to tackle the problem fully, thus, we propose to approach the problem by investigating rhetorical strategies employed by individuals in polarising discussions online. To this end, we develop multi-topic and multi-platform corpora with manual annotation of appeals to ethos and pathos, two modes of persuasion in Aristotelian rhetoric. It can be employed for training language models to advance the study of communication strategies online on a large scale. With the use of computational methods, our corpora allows an investigation of recurring patterns in polarising exchanges across topics of discussion and media platforms, and conduct both quantitative and qualitative analyses of language structures leading to and engaged in polarisation. 
\end{abstract}


\section{Introduction} \label{sec:intro}

\begin{figure}[h]
    \begin{center}
    \includegraphics[scale=0.35]{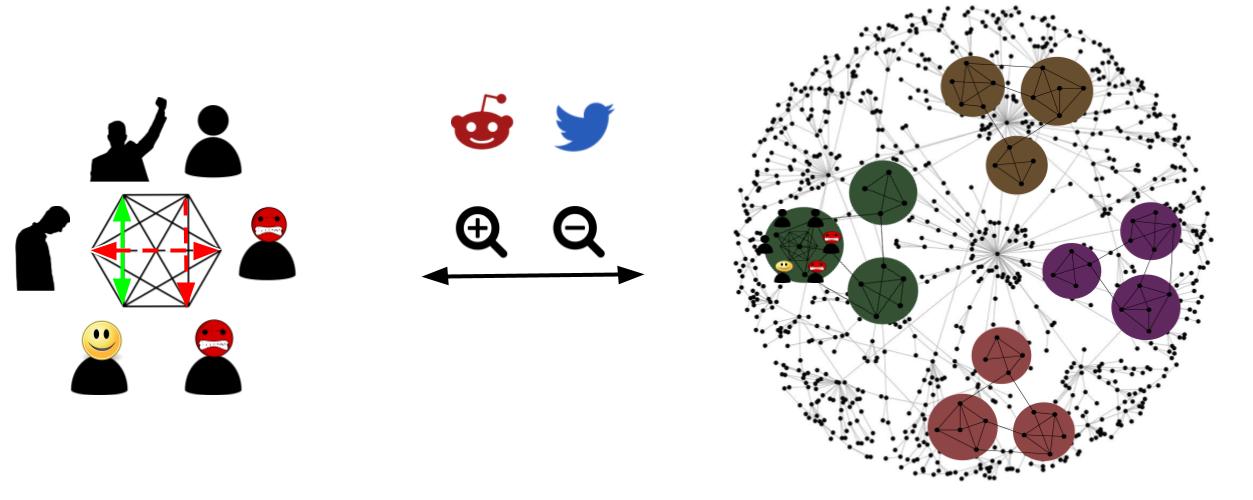}
    \caption{Our approach to the study of polarisation in group discussions using rhetorical strategies of appealing to ethos and pathos. Appealing to ethos is marked by a silhouette with a raised hand and a silhouette with a downward head tilt in cases of support and attack, respectively. Pathos is illustrated with the use of face emojis with either a positive (a smiling face) or negative (an angry face) expression.} 
    \label{fig:introfig} 
    \end{center}
\end{figure}

In a polarising environment individuals try to using both well structured argumentation as well as ad hominem fallacies \cite{kennedy1993aristotle}. 
Existing approaches to the study of polarisation online involve analyses of emotional language using hate speech detection and sentiment analysis tools, which in our view are insufficient for a robust research on polarisation in online spaces. 
Our language resources allow us to tackle the problem from a new perspective of \textbf{rhetorical appeals} to ethos and pathos, focusing on the hearer's perspective, complementing existing accounts.

Hate speech detection helps in removing profanity from a public debate, however, personal attacks involve also cases without explicit usage of swear words when referring to, for example, gender stereotypes\footnote{See a response to a female video game critic on Twitter: \url{https://femfreq.tumblr.com/post/52673540142/twitter-vs-female-protagonists-in-video-games}}. 
Then, sentiment analysis can indicate an expression of strong negative emotions of a speaker, however, the perspective of the hearer is left understudied.
Rhetorical strategies of appealing to ethos and pathos account for this gap and, thus, offer a more comprehensive  picture of communication practices in cases of conflict and polarisation in society.

The key requirement for the corpus design is to allow the conduct of large-scale analyses of polarising discussions online. 
\cite{lamm1978group} define \textbf{polarisation} as a phenomenon of strengthening individual attitudes through group interactions: individuals strive to protect the status of their in-group (in-group favouritism) while holding negative feelings towards the out-group (out-group hostility). Such defined polarisation is both issue-driven and affective \cite{bliuc2021online}. 

Using rhetorical appeals to ethos and pathos we account for both aspects of polarisation, and thus, propose a unique methodology of approaching the problem. 
Our approach here is presented in Figure \ref{fig:introfig}.
Our language resources are suitable for obtaining large-scale patterns of polarising behaviour by developing automatic tools for its processing, following recent works in the area such as computational approaches to pathos \cite{konatpathosargumentation}. 
Our resources also enable linguistic analyses of the language of polarisation, following recent works \cite{TNEPolarisingreportbook}. In this way, we can zoom in and out to investigate the phenomenon on different levels of quantitative and qualitative analysis.

In this paper, we introduce \textbf{multi-topic} and \textbf{multi-platform} \textbf{corpora} annotated with rhetorical strategies of appealing to ethos and pathos, two modes of persuasion distinguished by Aristotle in Rhetoric \cite{AristotleRhet}. Our language resources consist of online discussions on 3 polarising topics and are collected from 2 social media platforms in order to allow studying persuasive behaviour across online communities with different socio-demographic indicators \cite{cinelli2021echo}. 
Thus, our corpora help us obtain the knowledge on polarising patterns depending on the topic of discussion or the online platform as well as universal tendencies of polarising behaviour. 


Appeals to ethos refer to \textbf{the character of a speaker}, that is, her/his credibility. Such references can be favourable or unfavourable, that is one's ethos can be supported or attacked by others, respectively. Individuals, groups or organisations that participate in a discussion or are just a third party that is mentioned in the exchanges could be targets of such appeals.

\ex.  \label{ex:exx1} Xatencio00: \textit{\underline{Lie} number one goes to \underline{Hillary} about women getting paid less for the same work.}

Xatencio00 attacks the ethos of Hillary Clinton in Ex. \ref{ex:exx1}, referring to what candidate Clinton said during live debate and what the user regards as lying. There is a clear indication of a target as well as a negative connotation associated with her. 


\ex.  \label{ex:exx5} Boby: \textit{"You will own nothing and be happy"}

A sole reference to an entity is not enough to establish the presence of an ethotic appeal. Even though an entity (`you') is mentioned in Ex. \ref{ex:exx5}, s/he is neither supported nor attacked by the speaker. Thus, we consider such sentences as not appealing to ethos.

Pathos is the second element from the Aristotelian triad we provide resources for in this work. Persuasive power of pathotic appeals lies in their ability to \textbf{induce emotions in the hearers}. Thus, it focuses on affective states of the hearer in contrast to a commonly employed perspective of a speaker and emotions expressed in text content. 
\\

\ex.  \label{ex:exx3} MelsyJ: \textit{At least i do not \underline{bully disabled people} while displaying support for another country that has nothing to do with me.} 

Example \ref{ex:exx3} is a case of appealing to negative pathos of the hearers for its persuasive goals. It evokes negative emotions in the hearers; by an induction of empathy it can change their interpretation of arguments posed in a discussion. 

\ex.  \label{ex:exx2} Splenda: \textit{And, yes, \underline{California did us proud} today on the gas car phase-out, as Washington State did a few months ago! }

Finally, speakers can employ both rhetorical strategies in the same utterance.
In example \ref{ex:exx2} Splenda praises the ethos of the California state officials for their noble actions, also inducing positive emotions in the hearers. The exclamation mark at the end of the sentence in addition emphasises the emotional load of the message.

The paper is structured as follows. First, we present a methodology of a data collection. In Section \ref{sec:Relatedwork} we summarise existing resources for ethos and pathos in natural language. Section \ref{sec:ethos} presents annotation guidelines for ethos, results of the annotation and a discussion of challenging cases. Section \ref{sec:pathos} describes the annotation process of pathos appeals, summarises results of the annotation, and provides an analysis of the disagreement between annotators. Lastly, we show how our resources could be employed for the study of polarisation in social media debates in Section \ref{sec:Polarisationanalysis}, and conclude our work and findings in Section \ref{sec:Conclusions}.

\section{Data Collection} \label{sec:datacollection}
Our annotation of ethos and pathos extend previous works on logos in case of Reddit reactions to US 2016 presidential debates \cite{VisserKonat} and Reddit discussions on COVID vaccination. 
We also add 3 new large datasets annotated with selected rhetorical strategies: Twitter debates on COVID vaccination, Reddit debates on climate change, and Twitter debates on climate change.
The corpus is called \textbf{PolarIs} (PI) to indicate that it collects discussions on \textbf{Polarisation Issues}. It is available upon request via email. 
Our naming convention for these 5 datasets is presented in Table \ref{tab:lrnaming}. 
PI5 comprises reactions to live presidential debates, thus we did not collect an analogous dataset from Twitter because of the methodology of data collection in PI5 that would be challenging to reconstruct.

\begin{table}[h]
\caption{Language resources for \textit{Polarisation Issues} (\textbf{P}olar\textbf{I}s, PI) in  online group  discussions across 3 topics and 2 social media platforms.} \label{tab:lrnaming}
\centering
\begin{tabular}{|l|ll|}
\hline
\multicolumn{1}{|c|}{\cellcolor{lightgray} Topic / Platform} & Reddit & Twitter \\ \hline
 COVID-19 Vaccines & PI1    & PI2     \\ 
Climate Change & PI3    & PI4     \\ 
US2016 Elections & PI5    & x       \\ \hline
\end{tabular}

\end{table}

Two of our datasets (PI1 and PI3) are collected from specific groups on Reddit: r/conspiracy regarding the former, and r/climatechange and r/climateskeptics regarding the latter corpus, that is, largest discussion groups on conspiracy theories on vaccination and climate change, respectively. PI5 consists of reactions to 3 US 2016 presidential debates: the first Republican debate, the first Democratic debate, and the first general election debate. The detailed process of data collection for PI5 is presented in \cite{VisserKonat}. We retrieve this corpus from the AIFdb repository\footnote{\url{http://corpora.aifdb.org}}.

Then, collection of tweets in PI2 and PI4 is based on a keyword search. Regarding the former it is \#covid, \#vaccines, \#antivaxx, \#antivaccination, \#sarscov, \#phizer and \#vaccinated, and regarding the latter corpus it is \#climatechange, \#globalwarming, \#climatechangehoax, \#globalwarmingfraud and \#climatecrisis, following related works \cite{tyagi2020polarizing}. 
Here, we search for the occurrence of the keywords only in the first tweet that starts a given discussion, and then retrieve all replies to such tweet. Thus, our corpora comprise of social media discussions rather than single comments, which is rarely employed in related language resources.

\begin{table}[h]
 \caption{Corpora summary: its size in terms of the number of Words and Sent. (sentences), Source  (speakers) and Target (speakers) involved.} \label{tab:sizestats}
\begin{center}
\begin{tabular}{|lllll|}
    \hline
Corpus & Word&  Sent. & Source & Target \\ \hline
PI1   & 26,564    & 2,706  & 465 & 152        \\ 
PI2   & 27,136    & 2,488  & 499 & 192        \\ 

PI3    & 35,801    & 2,797  & 439 & 200        \\
PI4   & 35,742    & 3,334  & 1,050 & 268       \\ 

PI5 & 37,035    & 4,263  & 1,318 & 89        \\ \hline 

Total & 162,278   & 15,588 & 3,766 & 831 \\
\hline
    \end{tabular}
   
\end{center}        
\end{table}

\begin{figure}[h]
    \centering
    \includegraphics[width = 0.35\textwidth]{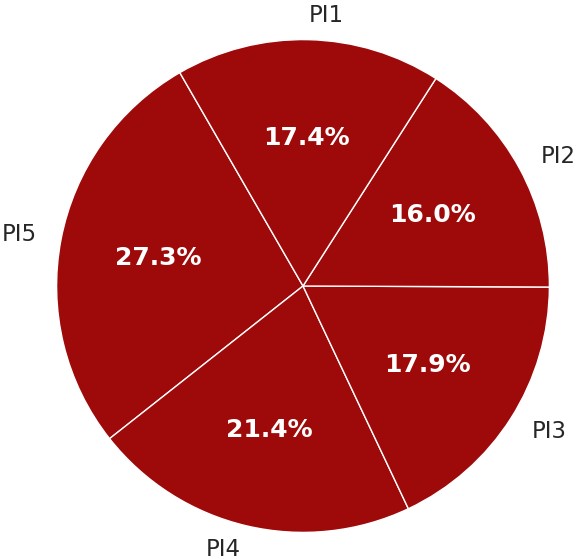}
    \caption{Percentage distribution (\%) of subcorpora size.}
    \label{fig:PICoverage}
\end{figure}

Table \ref{tab:sizestats} reports detailed statistics of our PolarIs corpora. In total it comprises 15,588 sentences annotated with appeals to ethos and pathos. 
We conduct automatic splitting of comments into sentences with SpaCy library in Python programming language. 
Fig. \ref{fig:PICoverage} displays a percentage distribution in terms of size (number of sentences) of individual datasets in our PolarIs corpus. 

In addition, we annotate our language resources with \textbf{reported speech}, that is, cases when one speaker reports what another speaker said, that is ``A said that B said x'', often enclosed in quotation marks. The phenomenon is illustrated in Ex. \ref{ex:exx5} and \ref{ex:exxerrorethos3}. Such cases are annotated as an additional layer. 
Reported speech in our language resources covers a small fraction of sentences though - it is between 0.6\% (PI2) and 4.0\% (PI5) of sentences, with the average inter-annotator agreement of 0.751 using Cohen's $\kappa$.

\section{Related Work} \label{sec:Relatedwork} 
While there exist many resources annotated with logos (appeals to reason) designed for argument mining tasks, the other two modes of persuasion are underexplored in computational linguistics. 
Resources that involve annotation of ethotic appeals cover either cases of self-referential ethos \cite{hidey2017analyzing} or data with formal language of political debates \cite{duthie2018deep,koszowy2022theory}. 
Regarding emotions, the majority of available resources involve the annotation of expressed emotions \cite{mohammad2012emotional,liew2016emotweet}. The idea of persuasive appeals to pathos that focus on emotions induced in the hearers is rarely implemented in computational linguistics. 
Thus, we aim to fill this gap by offering corpora annotated with both rhetorical strategies of appealing to ethos and pathos suitable for mining purposes and quantitative analysis.


\cite{duthie2018deep} provide a corpus of 90 transcripts of Hansard UK Parliamentary debates from the period of Margaret Thatcher as Prime Minister from 1979 to 1990. A sentence-level annotation of ethos is supplemented by target and speaker annotation. The authors observe 20\% of sentences constitute ethotic appeals.

Others \cite{habernal2018before} annotate types of ad hominem arguments, a form of ethotic attacks, and demonstrate difficulties associated with such annotation on a material from Reddit. \cite{habernal2018before} propose an annotation scheme of five sub-types of such arguments, namely abusive, tu quoque, circumstantial, bias and guilt by association. 
Results show abusive ad hominem to be the dominant types, with ambiguous interpretation of the other types and a high level of disagreement between raters. \cite{habernal2018before} observe that longer text spans such as social media posts often involve a mix of different categories, which is not taken into account by theoretical typologies. 

\cite{koszowy2022theory} in turn provide a fine-grained annotation of the Aristotelian concept of ethos, that is three elements of the speaker’s character: practical wisdom, moral virtue and goodwill, which can be either supported or attacked. \cite{koszowy2022theory} follow an iterative annotation process (agile corpus creation), which allows to achieve a suitable adaptation of theory to the practice of natural language communication and thus, cover complex cases of appeals to ethos. As a result, polymorphic ethos elements are introduced to account for cases of an overlap between three basic categories. Wisdom is the ethos element that is most frequently appealed to in ethotic supports (64.5\%), and virtue is a dominant category in attacks (34.8\%). Polymorphic categories of ethos elements constitute 19\% of cases.



Regarding appeals to pathos, \cite{mohammad2018wikiart} annotate induced emotions on a material of art content. Predominantly positive experiences such as love or optimism are induced by art. Such annotation in natural language is provided in the EMOBANK corpus \cite{buechel2022emobank}. It offers a bi-representational annotation of emotions from both perspectives of the speakers and the hearers on a material of blog posts. The annotation is conducted on a sentence level with the use of a dimensional account on the nature of emotions. Reliability of the annotation using Pearson’s correlation coefficient yields a satisfying agreement of 0.6 for both perspectives of the speaker and the hearer. We offer the annotation of appeals to positive and negative pathos on a material from social media. 

\cite{konatpathosargumentation} provide a corpus of political debates annotated with pathotic argumentation schemes. The authors find pathos-eliciting argumentation to be a commonly employed rhetorical strategy by US and Polish presidential candidates. Political actors appeal there to both positive and negative pathos of the hearers using mostly the argument from consequences.

The annotation of both rhetorical strategies of appealing to ethos and pathos is provided in \cite{hidey2017analyzing} on a material of argumentative texts. However, the researchers focus on cases of self-referential ethos, and observe only a small fraction of data to involve establishment of one's own ethos (4\% of cases). We propose corpora with appeals to the character of others, and observe a higher frequency of occurrence of such appeals to ethos. We observe, however, a similar density of pathos appeals in our corpora (29\% and 31\%).


\section{Ethos on Social Media} \label{sec:ethos}

\begin{table}[h]
 \caption{Summary of the ethos annotation: (1) Ethos I: distinction of sentences that contain ethos appeals (Ethos) and does not contain ethos (Non-Ethos). (2) Ethos II: classification between appeals to positive ethos (Support) and negative ethos (Attack). (3) Inter-annotator agreement (IAA) for Ethos I and Ethos II, and recognition of Target. } \label{tab:ethosstats} 
\begin{center}
\begin{tabular}{|l|cccc|cccc|ccc|}
\hline

\multicolumn{1}{|l|}{Corpus} &  \multicolumn{4}{c|}{Ethos I} &  \multicolumn{4}{c|}{Ethos II} &  \multicolumn{3}{c|}{IAA} \\ \cline{2-12}

& \multicolumn{2}{c|}{Ethos} & \multicolumn{2}{c|}{Non-Ethos} & \multicolumn{2}{c|}{Support} & \multicolumn{2}{c|}{Attack} & \multicolumn{3}{c|}{ $\kappa$}  \\ \cline{2-12}
{} & \# & \% & \# & \% & \# & \% & \# & \% & I & II &  Target  \\
\hline

Baseline  & 638  & 17.5 & 3,007 & 82.5 & 169 & 4.6 & 469 & 12.9 & 0.67  & 1.0  & 0.93 \\  \hline 
PI1  & 499  & 18.4  & 2,207 & 81.6 & 59 & 2.2$^*$ & 440 & 16.2 & 0.55 & 0.95 & 0.41 \\ 
PI2 & 587  & 16.9  & 1,901 & 83.1 & 120 & 3.5 & 467      & 13.4 & 0.64 & 0.89 & 0.76 \\
PI3  & 457  & 16.2  & 2,340 & 83.8 & 73 & 2.6$^*$ & 384      & 13.6 & 0.62 & 0.83 &  0.45\\ 
PI4  & 582  & 17.5 & 2,752 & 82.5 & 110 & 3.3 & 472 & 14.2 & 0.75 & 0.94 & 0.80 \\ 
PI5  & 1,339 & 31.4$^*$  & 2,924 & 68.6 & 492 & 11.5$^*$ & 847 & 19.9 & 0.79 & 0.80 & 0.83 \\ \hline
PI Avg/Total & 3,464 & 20.1 $\pm$6.4  & 12,124 & 79.9 & 854      & 4.6 $\pm$3.9 & 2,610     & 15.5 $\pm$2.7 & 0.67  & 0.88 & 0.65 \\ \hline 
    \end{tabular} 
   
\end{center} 
\end{table}

\paragraph{Annotation scheme.}
In the first step of the annotation process, a rater establishes the presence of ethotic appeal if both (i) a sentence mentions an entity; and (ii) s/he is mentioned in a favourable or unfavourable manner. In the second step, the annotator decides whether the reference is favourable (constitutes an ethotic support) or unfavourable (ethotic attack).
This study is built upon annotation of ethos in the UK parliament \cite{duthie2018deep}, adapted to the specificity of ethotic interactions in social media. 

Besides conversation-openers, posts are usually written as replies to previous posts. Thus, reading them is necessary to understand the meaning of the whole conversation, especially when resolving targets of ethotic appeals. Sentences are automatically extracted from social media posts and comprise individual units of annotation, therefore, annotators are encouraged to read whole posts that these text units are a part of in order to understand the meaning of individual sentences.

\paragraph{Results.}
Summary of the annotation results is presented in Table \ref{tab:ethosstats}. The density of ethotic appeals ranges from 16.2\% (PI3) to 31.4\% (PI5) in our corpora of social media discussions. We observe similar densities across four corpora. The results correspond to the previous findings on ethos appeals in the UK parliamentary debates  \cite{duthie2018deep}, which are considered baseline in Table \ref{tab:ethosstats}. However, PI5 constitutes an exception -- it is over 56\% more dense in ethotic appeals than the average, and over 84\% than the baseline. We suspect it is the result of a topic of discussion as the US political landscape is divided into two major groups. Therefore, support of the fellow members and attacks on the opponents might occur more naturally and thus, more frequently compared to the other two topics that are related to political affiliations, however, do not correspond to them directly \cite{merkley2018party,hegland2022partisan}. 

Ethos supports range from 2.2\% (PI1) to 11.5\% (PI5) and ethos attacks from 13.1\% (PI2) to 19.9\% (PI5). Thus, people more frequently perform attacks rather than praise others. We observe the density of ethotic supports to vary substantially across corpora -- in 3 cases (PI1, PI3, PI5) it is over 35\% difference over baseline and average values. Ethotic supports tend to be less common on Reddit compared to Twitter, except PI5 which consists of discussions on presidential elections.

\paragraph{Inter-annotator agreement.}
Part of the corpora was annotated in a traditional way, i.e. by an expert annotator (PI1, PI2, PI5). Regarding the rest, we implemented a time-efficient procedure ensuring high quality of the annotated resources. That is, two non-expert raters annotated the full corpora; then the disagreements between them were resolved by an expert annotator (super-annotator). 
Inter annotator agreement (IAA) coefficients reported in Table \ref{tab:ethosstats} confirm reliability of both approaches to the annotation process. 
 
We calculate IAA using Cohen's $\kappa$ and follow a two-step procedure introduced in \cite{duthie2016mining}: first, we compute coefficients for the detection of ethotic appeals (Ethos I); second, for the classification of the polarity of ethotic appeals with support versus attack labels (Ethos II).
Inter-annotator agreement for ethos is adequate. Cohen’s kappa for distinguishing between ethos and non-ethos ranges between 0.55 (PI1) to 0.79 (PI5), and for distinguishing between ethos support and attack ranges between 0.80 (PI5) to 0.95 (PI1). 
Therefore, we conclude it is more challenging to assert an appeal to ethos in a sentence than recognise its polarity -- if two raters agree there is an appeal to ethos, it is very straightforward to recognise its polarity (support versus attack). These results accord with the previous findings in the corpora of parliamentary debates \cite{duthie2016mining,duthie2018deep}.

The annotation of targets is substantially more demanding than recognition of ethotic appeals, as there is no finite set of names or labels to choose from (see Table \ref{tab:ethosstats}). Nonetheless, we note substantial agreement in our corpora. 
Also, we observe the annotation of targets to be more challenging on Reddit compared to Twitter. Regarding the latter platform, replies to other tweets start with the names of user accounts (marked with the ``@'' sign), so that annotators have a straightforward task of resolving, e.g., ``you'' pronouns in sentences. In the case of Reddit, the task is more demanding as annotators need to track the discussion thread to resolve such pronouns, although highest IAA coefficients are observed in PI5. 

We suspect this difference arises from the number of entities targeted in PI5 compared to the others. Discussion on US 2016 presidential elections has a well defined set of actors that could become targets of ethotic appeals (presidential candidates and US politicians). As a result, the identification of targets of appeals is the least challenging here.

\paragraph{Error analysis.}
Challenging cases in the annotation of ethos involve linguistic structures such as questions, irony/sarcasm and reported speech. Appeals to ethos (usually attacks) seem to be softened or hidden in questions and thus, might be left unnoticed by non-expert annotators. 



\ex. \label{ex:exxerrorethos2} jonny4: \textit{I’m sure \underline{they} are fully and \underline{happily aware of the consequences!}}

Cases such as Ex. \ref{ex:exxerrorethos2} pose a challenge for human annotators as the use of irony or sarcasm is implicitly stated in a sentence, thus, not always easily recognised by raters. 

\ex. \label{ex:exxerrorethos3} Turrubul: \textit{"and he does a great job"}

Lastly, we note some cases of reported speech, where speakers quote the words articulated by other people as in Ex. \ref{ex:exxerrorethos3}. However, if a speaker just quotes an (un)favourable reference to some entity, we do not regard such cases as appeals to ethos. We observe it is challenging to identify such sentences and distinguish actual cases of ethotic appeals from those just quoted by speakers with a neutral connotation as in Ex. \ref{ex:exxerrorethos3} one annotator marks ethotic attack and the other regards the sentence as not appealing to ethos, which is correct in this case.

\section{Pathos on Social Media} \label{sec:pathos}

\begin{table}[h]
\caption{Summary of pathos annotation: (1) Pathos I: identification of sentences that contain appeals to pathos (Pathos) and does not contain pathos (Non-Pathos). (2) Pathos II: distinction between appeals to positive pathos (Positive) and negative pathos (Negative). (3) Inter-annotator agreement (IAA) for Pathos I and Pathos II.} \label{tab:pathosstats}
\begin{center}
    \begin{tabular}{|l|cccc|cccc|cc|}
    \hline
    {Corpus} & \multicolumn{4}{c|}{Pathos I}  & \multicolumn{4}{c|}{Pathos II} &  \multicolumn{2}{c|}{IAA} \\ \cline{2-11}
         {} & \multicolumn{2}{c|}{Pathos}  & \multicolumn{2}{c|}{Non-Pathos} & \multicolumn{2}{c|}{Positive} & \multicolumn{2}{c|}{Negative} & \multicolumn{2}{c|}{ $\kappa$}  \\ \cline{2-11}
{} & \# & \% & \# & \% & \# & \% & \# & \% & I & II \\ 
\hline

PI1       & 805  & 29.7 & 1,901 & 70.3 & 152      & 5.6 & 653      & 24.1 & 0.39 & 0.78 \\ 
PI2       & 886  & 24.3 & 1,602 & 75.7 & 195      & 5.3 & 692      & 19.0 & 0.33 & 0.65  \\ 
PI3       & 862  & 30.8  & 1,935 & 69.2 & 107      & 3.8 & 755      & 27.0 & 0.33 & 0.65 \\ 
PI4       & 1,186 & 35.6 & 2,148 & 64.4 & 251   & 7.5$^*$ & 935      & 28.1 & 0.33 & 0.65 \\  
PI5       & 1,484 & 34.8  & 2,779 & 65.2 & 190      & 4.5             & 1,294     & 30.4 & 0.32 & 0.82 \\ \hline  
Avg/Total & 5,223 & 31.0 $\pm$4.5 & 10,365 &  69.0 & 895      & 5.3 $\pm$1.4 & 4,329 & 25.7 $\pm$4.4 & 0.34  & 0.71   \\
        \hline
    \end{tabular}
\end{center}        
\end{table}

\paragraph{Annotation scheme.}
The distinction between perspectives of the speaker (expressed emotions) and the hearers (induced emotions) of a message was acknowledged in recent works involving annotation of pathos-appealing argument schemes \cite{konatpathosargumentation}, lexicons of emotion-eliciting words \cite{wierzba2021emotion}, or bi-representational corpus with valence-arousal-dominance (VAD) representation format \cite{buechel2022emobank}. 
To the best of our knowledge, our resources are the first large database of patothic appeals in natural language, fully annotated by human raters on a sentence level and data from social media. Annotation of pathos was conducted by students of cognitive linguistic course, trained by one of the authors in-house ensuring supervision and high quality of annotation.

We design a simple two-step procedure for the annotation of pathotic appeals in natural language. First, we focus on the recognition of the speaker's intention of a rhetorical gain coming from using emotion-eliciting language. Second, we determine whether the emotional state the speaker attempts to induce in the hearers is positive or negative based on linguistic markers in text. These cues comprise single words, phrases, and rhetorical figures such as metaphors and irony.

\paragraph{Results.}
The density of pathos ranges from 24.3\% (PI2) to 35.6\% (PI4), reported in Table \ref{tab:pathosstats}. Thus, in all cases as well as on average we observe a higher density of appeals to pathos than ethos in our data. Ethos is observed in 20.1\% of sentences, including 4.6\% supports and 15.5\% attacks. Appeals to pathos regard 31.0\% of sentences, including 5.3\% positive and 25.7\% negative, on average. Therefore, discussions in our corpus are 54\% more dense in appealing to emotions of the hearers than the character of persons. 

These results accord with the previous findings that pathos was more frequently employed by US 2012 presidential candidates, Obama and Romney, in their social media activity \cite{bronstein2013like}. In a political genre pathos-appealing argument schemes seem to be also a popular rhetorical strategy in presidential debates  \cite{konatpathosargumentation}. In this corpus it constitutes 52\% of all arguments, and emotion-eliciting language can be found in over 90\% of arguments. 
The density of pathotic appeals is also more stable than the density of appeals to ethos as standard deviations are equal to 4.5\% and 6.4\%, respectively. Since the latter strategy depends on the actors related to a specific topic of discussion, its density is more varied across corpora. 
We suspect appeals to pathos to be the rhetorical strategy more straightforward to employ than appeals to ethos, thus, they are observed more frequently in our corpora.

Positive pathos ranges from 3.8\% (PI3) to 7.5\% (PI4) and negative pathos from 18.9\% (PI2) to 30.4\% (PI5). Thus, we observe a similar trend of the dominance of a negative category of pathos appeals over a positive one as in the case of appeals to ethos. Moreover, PI2 and PI5 are the two datasets least and most dense in negative pathos and ethotic attacks at the same time. 
The density of appeals to positive and negative pathos is similar across datasets; only in PI4 we observe over 40\% higher density of positive appeals compared to the average value (7.5\% versus 5.3\%). 

The distribution of positive pathos and ethotic supports are on a similar level (around 5\%), however appeals to negative pathos are substantially more prevalent than ethotic attacks (26\% vs. 15\%). Thus, the difference between the density of pathos and ethos (31\% vs. 20\%) arises because of the higher amount of negative appeals of the former rhetorical strategy.

\paragraph{Inter-annotator agreement.}
Reliability of pathos annotation is calculated by analogy to ethos, i.e., in a two step manner, and reported in Table \ref{tab:pathosstats}. 
Inter-annotator agreement is adequate. However, we note the annotation of pathos to be more challenging than the annotation of ethos.
We observe lower values of the $\kappa$ score for Pathos I compared to Ethos I, and similar values of the annotator agreement for the classification of polarity of both rhetorical appeals to ethos and pathos (Pathos II and Ethos II).
We suspect lower IAA coefficients for the detection of pathos appeals come from the higher degree of subjectivity of the former phenomenon. Reference to an entity in the case of ethos is explicitly marked in text, and thus it is a more objective task to assert an appeal to ethos and as a result more probable that two annotators agree in the annotation process. 

Emotions are inherently subjective phenomena, and asking an annotator to immerse her/himself in the recipient's perspective to understand pathotic effect proved to be a challenging task. We tried to account for this fact by the employment of higher number of annotators for the pathos annotation compared to the ethos annotation. However our results are promising, as they show that at least some level of intersubjectivity is possible in the annotation of pathos. 


\paragraph{Error analysis.}
Challenging cases in pathos annotation involve similar language phenomena as in the case of ethos, that is, ironic utterances and questions as well as very short sentences. In addition we note a difficulty in distinguishing expressed versus evoked emotions for our annotators as well as polymorphic cases of appealing to both positive and negative pathos depending on the perception of an appeal by the hearer. 

\ex. \label{ex:exxerrorpathos1} SherylBR549: \textit{Poor us!} 

The use of irony or sarcasm makes it difficult to recognise intentions of a speaker, as it can both signal the opinion of the speaker on an issue as well as evoke a certain picture of the situation in the hearers together with its emotional load. Such sentences can evoke both positive and negative emotional states in the hearers by an induction of empathy and resentment, respectively. Ex. \ref{ex:exxerrorpathos1} thus involve a polymorphic case of appealing both to positive and negative pathos. Such cases could be implemented in future works in a new annotation scheme.


\ex. \label{ex:exxerrorpathos2} L46: \textit{I hate it that they essentially use this as motivation to avoid any mitigations.}

In Ex. \ref{ex:exxerrorpathos2} annotators confuse the concepts of expressed and evoked emotions. Here, the user expresses her/his opinion and uses words with negative connotation, which might be the reason we observe a disagreement in this case.





\section{Polarisation Issues}  \label{sec:Polarisationanalysis}

\begin{figure}[h]
    \centering
    \includegraphics[width=1\textwidth]{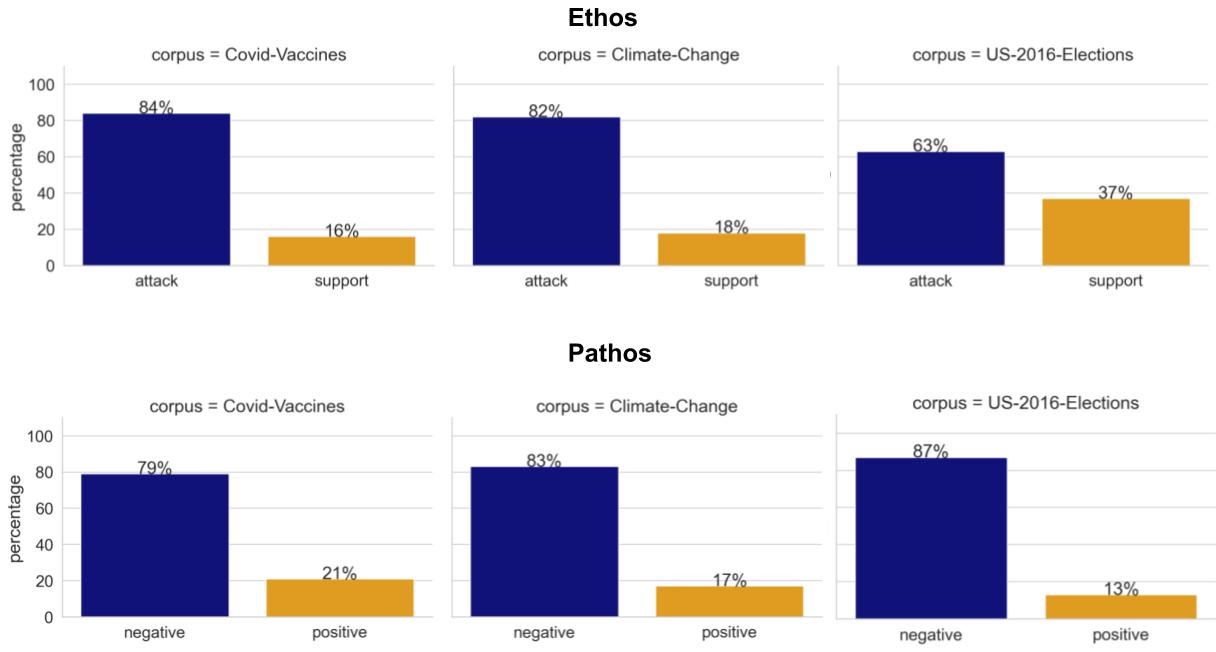} \caption{Negative versus positive categories of ethos and pathos appeals across topics of discussion. Ethos is presented at the top, and pathos at the bottom.}
    \label{fig:ethpanprop}
\end{figure}

The design of our corpora allows us to analyse both general patterns of polarising behaviour and specificities of particular media platforms and topics of discussion in their polarising tendency. 
Ethotic attacks are roughly 4 times more prevalent than ethotic supports, however discussion on US elections is characterised by a ratio of 2:1 of attacks to supports.
We observe politics-centred debates to be more balanced in ethotic appeals (the ratio of ethotic attacks to supports) than discussions on politics-related topics of COVID vaccination and climate change (see Fig. \ref{fig:ethpanprop}). 

We suspect these findings can be explained by a higher degree of polarisation of society in a politics-centred debate. In a highly polarised environment the public fragments into groups with separate authorities and enemies each group supports and attacks. In a less polarised situation, the public has a common set of enemies they collectively attack, thus, the number of ethotic attacks might increase. As a result, the ratio of attacks to supports is more imbalanced. 

We observe a similar trend for pathos in terms of the ratio of negative to positive appeals (4:1). However regarding the topic of US elections, we note negative appeals to be over 6 times more prevalent than positive appeals (see Fig. \ref{fig:ethpanprop}). Thus, contrary to the ethos annotation, we notice the US elections topic to be the most imbalanced in terms of the two categories of appeals to pathos. These results need further investigation which we plan to conduct in collaboration with social scientists.

We can also distinguish a platform dependent usage of rhetorical strategies -- praising the ethos of others is more common for the Reddit community in our corpora compared to Twitter, however, appealing to positive pathos is more common on Twitter than Reddit. 
We suspect these results arise because of platform peculiarities. 
On Reddit, there are well defined discussion groups (subreddits) designed around a common view on the issue such as climate change. On Twitter, a discussion is started by a single tweet and everyone is free to join the debate. 
Furthermore, ethotic supports are often performed towards fellows (that is, members of the in-group), who are easily identified on Reddit because of a clear demarcation of a group membership, whereas Twitter lacks such structures. 
On the other hand, positive appeals to pathos might be a more straightforward strategy to employ than ethos and thus, it is employed in a higher proportion on Twitter than Reddit to strike up a conversation with strangers and form such communities of like-minded individuals.

\begin{figure}[h]
    \centering
    \includegraphics[width=0.65\columnwidth]{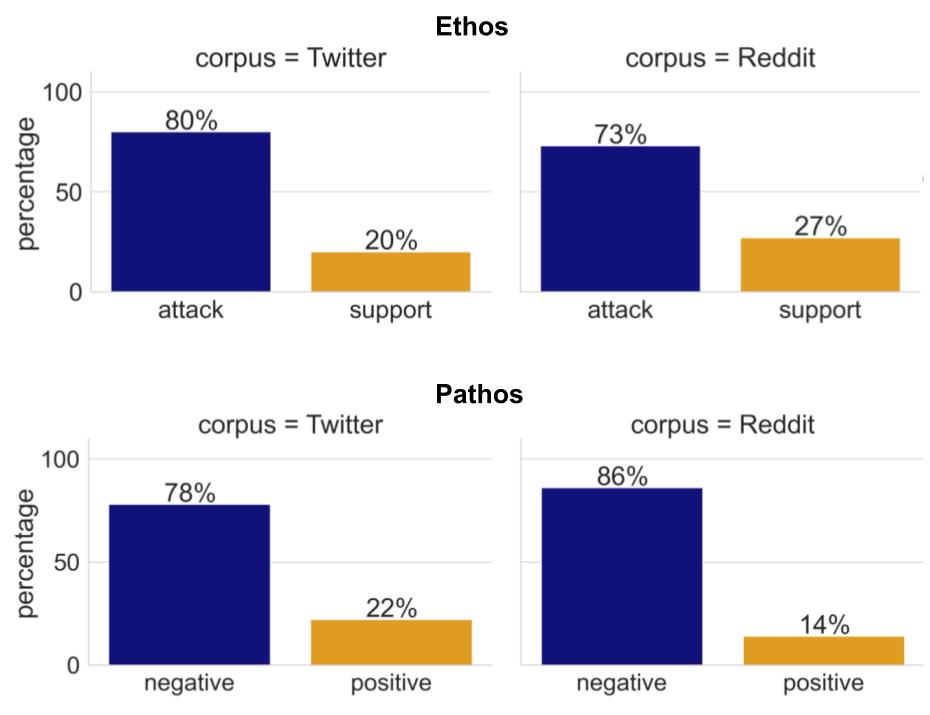}
    \caption{Negative versus positive appeals to ethos and pathos across social media platforms of Reddit and Twitter.}
    \label{fig:ethpanpropplatform}
\end{figure}

\section{Conclusions} \label{sec:Conclusions}
We propose a new approach to the study of polarisation of society, complementing existing methods and tools in computational linguistics. To this end, we present a large dataset of online discussions on polarising issues, annotated with selected rhetorical strategies from Aristotelian rhetoric. Our language resources thus fill the gap on underexplored areas of appealing to ethos and pathos, extending previous corpora with logos and studies on argumentation.  
Our approach also extends well-known techniques of hate speech detection and sentiment analysis, thus, enabling the analysis of both perspectives of the speaker and the hearer in polarising discussions. A multi-topic and multi-platform design allows further investigation of both universal as well as topic and platform dependent patterns of polarising behaviour. Here, we observe the usage of both rhetorical strategies of ethos and pathos is higher on Reddit than Twitter.

\nocite{*}
\section{Acknowledgements}
The work reported in this paper was supported in part by CHIST-ERA under grant 2022/04/Y/ST6/00001, in part by the Polish National Science Centre (NCN) under grant 2020/39/I/HS1/02861, in part by the European Union’s Horizon 2020 research and innovation programme under the Marie Skłodowska-Curie Grant Agreement No. 860621, and in part by VW foundation (VolkswagenStiftung) under grant 98 542.


\begin{thebibliography}{30}


\bibitem{duthie2018deep}
Rory Duthie and Katarzyna Budzynska.
A deep modular rnn approach for ethos mining.
In \emph{IJCAI}, pages 4041--4047, 2018.

\bibitem{konatpathosargumentation}
Barbara Konat, Ewelina Gajewska, and Wiktoria Rossa.
Pathos in natural language argumentation: emotional appeals and reactions.
\emph{Argumentation}, 20xx.
In press.

\bibitem{kennedy1993aristotle}
George~A Kennedy.
Aristotle" on rhetoric": a theory of civic discourse.
\emph{Philosophy and Rhetoric}, 26\penalty0 (4), 1993.

\bibitem{lamm1978group}
Helmut Lamm and David~G Myers.
Group-induced polarization of attitudes and behavior.
In \emph{Advances in experimental social psychology}, volume~11, pages 145--195. Elsevier, 1978.

\bibitem{bliuc2021online}
Ana-Maria Bliuc, Ayoub Bouguettaya, and Kallam~D Felise.
Online intergroup polarization across political fault lines: An integrative review.
\emph{Frontiers in Psychology}, 12:\penalty0 641215, 2021.

\bibitem{TNEPolarisingreportbook}
Ewelina Gajewska, Katarzyna Budzynska, Barbara Konat, and Marcin Koszowy, editors.
\emph{Linguistically Analysing Polarisation on Social Media}.
The New Ethos Reports, vol. 1, Warsaw Poland, Warsaw University of Technology, doi: 10.17388/WUT.2023.0001.AINS, March 2023.
\doi{10.17388/WUT.2023.0001.AINS}.

\bibitem{AristotleRhet}
Aristotle.
\emph{Rhetoric}.
Dover Publications, 2004.

\bibitem{cinelli2021echo}
Matteo Cinelli, Gianmarco De~Francisci~Morales, Alessandro Galeazzi, Walter Quattrociocchi, and Michele Starnini.
The echo chamber effect on social media.
\emph{Proceedings of the National Academy of Sciences}, 118\penalty0 (9):\penalty0 e2023301118, 2021.

\bibitem{VisserKonat}
Jacky Visser, Barbara Konat, Rory Duthie, Marcin Koszowy, Katarzyna Budzynska, and Chris Reed.
Argumentation in the 2016 {US} presidential elections: annotated corpora of television debates and social media reaction.
\emph{Language Resources and Evaluation}, 54\penalty0 (1):\penalty0 123--154, 2020.
\doi{10.1007/s10579-019-09446-8}.

\bibitem{tyagi2020polarizing}
Aman Tyagi, Matthew Babcock, Kathleen~M Carley, and Douglas~C Sicker.
Polarizing tweets on climate change.
In \emph{International Conference on Social Computing, Behavioral-Cultural Modeling and Prediction and Behavior Representation in Modeling and Simulation}, pages 107--117. Springer, 2020.

\bibitem{hidey2017analyzing}
Christopher Hidey, Elena Musi, Alyssa Hwang, Smaranda Muresan, and Kathy McKeown.
Analyzing the semantic types of claims and premises in an online persuasive forum.
In \emph{Proceedings of the 4th Workshop on Argument Mining}, pages 11--21, Copenhagen, Denmark, September 2017. Association for Computational Linguistics.
\doi{10.18653/v1/W17-5102}.

\bibitem{koszowy2022theory}
Marcin Koszowy, Katarzyna Budzynska, Martin Pereira-Fari{\~n}a, and Rory Duthie.
From theory of rhetoric to the practice of language use: The case of appeals to ethos elements.
\emph{Argumentation}, pages 1--27, 2022.

\bibitem{mohammad2012emotional}
Saif Mohammad.
\# {E}motional tweets.
In \emph{*SEM 2012: The First Joint Conference on Lexical and Computational Semantics--Volume 1: Proceedings of the main conference and the shared task, and Volume 2: Proceedings of the Sixth International Workshop on Semantic Evaluation (SemEval 2012)}, pages 246--255, 2012.

\bibitem{liew2016emotweet}
Jasy Suet~Yan Liew, Howard~R. Turtle, and Elizabeth~D. Liddy.
{E}mo{T}weet-28: A fine-grained emotion corpus for sentiment analysis.
In \emph{Proceedings of the Tenth International Conference on Language Resources and Evaluation ({LREC}'16)}, pages 1149--1156, Portoro{\v{z}}, Slovenia, May 2016. European Language Resources Association (ELRA).
URL \url{https://aclanthology.org/L16-1183}.

\bibitem{habernal2018before}
Ivan Habernal, Henning Wachsmuth, Iryna Gurevych, and Benno Stein.
Before name-calling: Dynamics and triggers of ad hominem fallacies in web argumentation.
In \emph{Proceedings of NAACL-HLT}, pages 386--396, 2018.

\bibitem{buechel2022emobank}
Sven Buechel and Udo Hahn.
{E}mo{B}ank: Studying the impact of annotation perspective and representation format on dimensional emotion analysis.
In \emph{Proceedings of the 15th Conference of the {E}uropean Chapter of the Association for Computational Linguistics: Volume 2, Short Papers}, pages 578--585. Association for Computational Linguistics, April 2017.

\bibitem{merkley2018party}
Eric Merkley and Dominik~A Stecula.
Party elites or manufactured doubt? the informational context of climate change polarization.
\emph{Science Communication}, 40\penalty0 (2):\penalty0 258--274, 2018.

\bibitem{hegland2022partisan}
Austin Hegland, Annie~Li Zhang, Brianna Zichettella, and Josh Pasek.
A partisan pandemic: how covid-19 was primed for polarization.
\emph{The ANNALS of the American Academy of Political and Social Science}, 700\penalty0 (1):\penalty0 55--72, 2022.

\bibitem{duthie2016mining}
Rory Duthie, Katarzyna Budzynska, and Chris Reed.
Mining ethos in political debate.
In \emph{Computational Models of Argument: Proceedings from the Sixth International Conference on Computational Models of Argument (COMMA)}, pages 299--310. IOS Press, 2016.

\bibitem{wierzba2021emotion}
Ma{\l}gorzata Wierzba, Monika Riegel, Jan Koco{\'n}, Piotr Mi{\l}kowski, Arkadiusz Janz, Katarzyna Klessa, Konrad Juszczyk, Barbara Konat, Damian Grimling, Maciej Piasecki, et~al.
Emotion norms for 6000 {Polish} word meanings with a direct mapping to the {Polish} wordnet.
\emph{Behavior Research Methods}, pages 1--16, 2021.

\bibitem{bronstein2013like}
Jenny Bronstein.
Like me! analyzing the 2012 presidential candidates’ facebook pages.
\emph{Online Information Review}, 37\penalty0 (2):\penalty0 173--192, 2013.

\bibitem{al2016arabic}
Ayman Al~Zaatari, Rim El~Ballouli, Shady ELbassouni, Wassim El-Hajj, Hazem Hajj, Khaled Shaban, Nizar Habash, and Emad Yahya.
Arabic corpora for credibility analysis.
In \emph{Proceedings of the Tenth International Conference on Language Resources and Evaluation (LREC'16)}, pages 4396--4401, 2016.

\bibitem{walton2013scare}
Douglas Walton.
\emph{Scare tactics: Arguments that appeal to fear and threats}, volume~3.
Springer Science \& Business Media, 2013.

\bibitem{scienespecialissue23}
Holden Thorp.
Social media and elections {[Special} issue].
\emph{Science}, 381\penalty0 (6656), 2023.

\bibitem{budzynska2011speech}
Katarzyna Budzynska and Chris Reed.
Speech acts of argumentation: Inference anchors and peripheral cues in dialogue.
In \emph{Computational models of natural argument: papers from the 2011 {AAAI} Workshop}, pages 3--10. {AAAI Press}, 2011.

\bibitem{pauli2022modelling}
Amalie Pauli, Leon Derczynski, and Ira Assent.
Modelling persuasion through misuse of rhetorical appeals.
In \emph{Proceedings of the Second Workshop on NLP for Positive Impact (NLP4PI)}, pages 89--100, 2022.

\bibitem{mohammad2018wikiart}
Saif Mohammad and Svetlana Kiritchenko.
{W}iki{A}rt emotions: An annotated dataset of emotions evoked by art.
In \emph{Proceedings of the Eleventh International Conference on Language Resources and Evaluation ({LREC} 2018)}, Miyazaki, Japan, May 2018. European Language Resources Association (ELRA).
URL \url{https://aclanthology.org/L18-1197}.

\bibitem{landis1977measurement}
J~Richard Landis and Gary~G Koch.
The measurement of observer agreement for categorical data.
\emph{biometrics}, pages 159--174, 1977.

\bibitem{patel2023dummy}
Utkarsh Patel, Animesh Mukherjee, and Mainack Mondal.
" dummy grandpa, do you know anything?": Identifying and characterizing ad hominem fallacy usage in the wild.
In \emph{Proceedings of the International AAAI Conference on Web and Social Media}, volume~17, pages 698--709, 2023.

\bibitem{rozin2001negativity}
Paul Rozin and Edward~B Royzman.
Negativity bias, negativity dominance, and contagion.
\emph{Personality and social psychology review}, 5\penalty0 (4):\penalty0 296--320, 2001.

\end{thebibliography}

\end{document}